\documentclass[conference]{IEEEtran}
\IEEEoverridecommandlockouts
\usepackage{algorithm}
\usepackage{cite}
\usepackage{amsmath,amssymb,amsfonts}
\usepackage{algorithmic}
\usepackage{graphicx}  
\usepackage{textcomp}
\usepackage{xcolor}     
\usepackage{booktabs}
\usepackage{multirow}
\usepackage{colortbl}
\usepackage{float}
\usepackage{tikz}
\usepackage{caption}   

\usepackage{hyperref}

\usepackage[capitalise]{cleveref}

\crefname{table}{Tab.}{Tabs.}
\Crefname{table}{Tab.}{Tabs.}

\def\BibTeX{{\rm B\kern-.05em{\sc i\kern-.025em b}\kern-.08em
    T\kern-.1667em\lower.7ex\hbox{E}\kern-.125emX}}

\newcommand{\circleone}[1]{%
    \resizebox{!}{0.8em}{%
        \tikz[baseline=(char.base)]{
            \node[shape=circle, fill=black, inner sep=0.8pt, text=white] (char) {#1};
        }%
    }%
}

\newcommand{\circletwo}[1]{%
    \resizebox{!}{0.8em}{%
        \tikz[baseline=(char.base)]{
            \node[shape=circle, fill=black, inner sep=0.8pt, text=white] (char) {#1};
        }%
    }%
}

\begin{document}

\title{VFACamou: View-Fused Adversarial Camouflage for Environment-Adaptive Physical Evasion}

\author{
 Shihui Yan$^{1,2*}$, Hu Liu$^{1*}$, Junyu Shi$^{2}$, Zihui Zhu$^{2}$, Ziqi Zhou$^{3}$, \\ Yufei Song$^{2}$, Youming Geng$^{5}$, Minghui Li$^{4}$, Shengshan Hu$^{2}$
 \\
 $^{1}$ State Key Laboratory of Intelligent Vehicle Safety Technology \\
 $^{2}$ School of Cyber Science and Engineering,  Huazhong University of Science and Technology   \\
 $^{3}$ School of Computer Science and Technology, 
 Huazhong University of Science and Technology \\
 $^{4}$ School of Software Engineering, Huazhong University of Science and Technology \\ 
 $^{5}$ Hebei Energy College of Vocation And Technology \\ 
 $*$ These authors contributed equally to this work\\

\footnotesize{\texttt{\{yanshihui,zhuzihui,string1,zhouziqi,yufei17,minghuili,hushengshan\}@hust.edu.cn}
}
\\
\footnotesize{\texttt{stevenliu\_vip@hotmail.com},  \texttt{andylongming@163.com}}
}

\twocolumn[{
\renewcommand\twocolumn[1][]{#1}
\maketitle
\begin{center}
    \centering
    \vspace*{-.5cm}
    \includegraphics[width=1.0\textwidth]{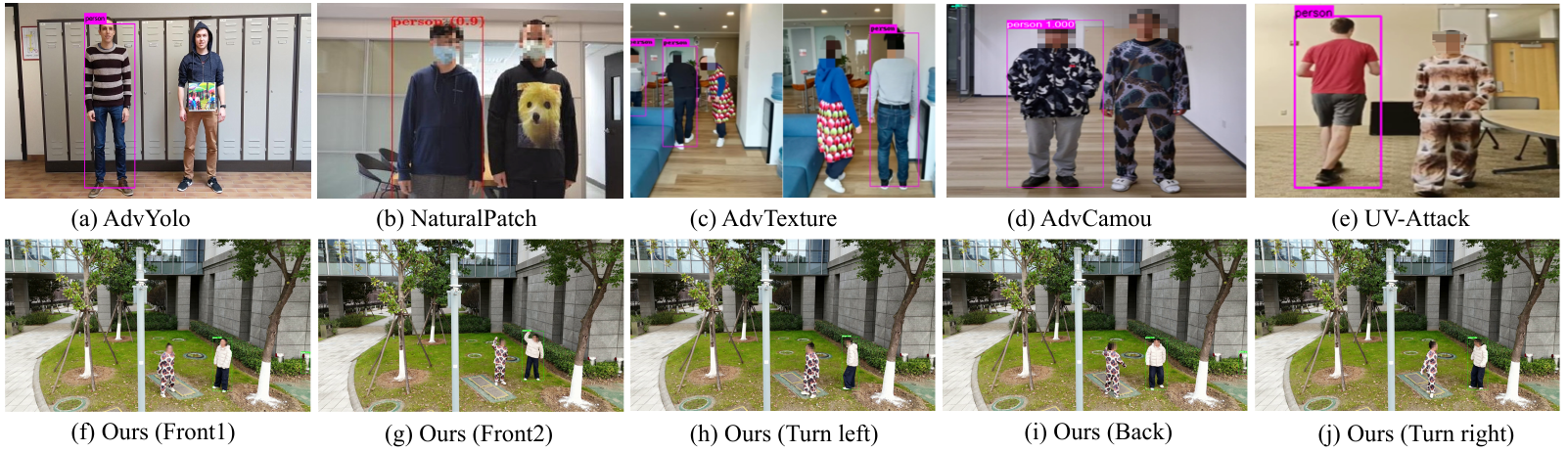}
    \captionof{figure}{
    Visualization of different person detector attacks.
    We compare our method with AdvYOLO~\cite{thys2019fooling}, NaturalPatch~\cite{hu2021naturalistic}, AdvTexture~\cite{hu2022adversarial}, AdvCamou~\cite{hu2023physically}, and UV-Attack~\cite{li2025uv}.
    Our adversarial textures generated using the ``trunk''  prompt demonstrate consistently high attack success rates across various viewpoints and poses.} 
\label{fig:UniMV}
\end{center}
}]

\begin{abstract}
Adversarial camouflage in the physical world remains highly challenging, particularly under UAV reconnaissance where targets undergo continuous geometric changes and extreme illumination variations. 
Existing methods either optimize 2D digital perturbations that fail to generalize to dynamic viewpoints or produce visually unnatural textures that cannot be deployed in real scenarios. 
Therefore, we propose VFACamou, an end-to-end framework for adversarial camouflage generation that automatically produces wearable adversarial patterns and maintains stable attack performance in real physical environments with changing viewpoints, poses, and lighting conditions.
Our method integrates UV-volume rendering with a diffusion-based texture generator, enabling consistent appearance under varying scales, poses, and lighting conditions. 
To ensure environmental realism, we propose an illumination–color consistency estimator that extracts dominant background attributes and guides a natural texture loss to align the generated UV texture with the surrounding environment.
A multi-scale dynamic training strategy further enhances robustness against viewpoint shifts and body deformation.
Extensive experiments across multiple mainstream detectors demonstrate that VFACamou achieves strong and stable physical attack performance while maintaining high perceptual naturalness, reducing human detection rates without introducing unnatural artifacts. 
\end{abstract}


\section{Introduction}
\label{sec:intro}
With the rapid development of deep learning, unmanned aerial vehicles equipped with intelligent vision systems play an increasingly important role in border patrol, area monitoring, and situational awareness\cite{tang2024surveyuav,ren2015fasterrcnn,du2018visdrone}. 
Recent studies show that modern object detectors face various security risks~\cite{zhou2025darkhash,hu2022badhash,wan2025mars,zhang2024detector,zhang2025test,wang2024trojanrobot,wang2024unlearnable,li2025detecting,wang2024eclipse,yu2025spa,xue2026towards} and are highly vulnerable to adversarial attacks, in which carefully crafted perturbations can mislead the detectors\cite{zhou2024darksam,zhou2025sam2,wang2025advedm,song2025segment,li2024transferable,song2025seg}. In military  scenarios, generating effective adversarial camouflage to shield personnel from automatic detection holds significant  importance\cite{advclip,zhou2024securely}.

Existing adversarial attacks exhibit notable limitations in UAV reconnaissance scenarios, as most studies concentrate on 2D image settings while neglecting real-world complexities\cite{wang2025breaking}. In particular, current methods lack robustness to geometric transformations arising from varying altitudes and viewpoints, which induce continuous changes in scale and perspective and thus degrade attack effectiveness. Moreover, they are insufficient in handling dynamic environmental conditions, especially severe illumination variations. Changes in lighting intensity, color temperature, and direction—caused by factors such as time, weather, and terrain—significantly alter the visual appearance of targets, making textures generated under constant-lighting assumptions difficult to maintain consistency in real-world applications\cite{liu2025lighting}.

Moreover, a prominent limitation of existing approaches lies in their limited naturalness. To maximize attack strength, many generated textures contain unnatural high-frequency artifacts or colors that are inconsistent with the surrounding environment\cite{li2025uv}. 
Such adversarial camouflage patterns are easily observed by humans and are therefore difficult to be deployed effectively in real-world scenarios.

To address these limitations, this work targets a central problem: generating adversarial camouflage that consistently deceives both deep neural networks and human observers in physical UAV reconnaissance environments. 
This problem presents two key challenges:
\circleone{1} \textit{Robustness in dynamic observation.}   The generated camouflage must maintain effective and stable adversarial attacks under continuous changes in scale and viewpoint caused by UAV motion.
\circletwo{2} \textit{Balancing adversarial effect and environmental fusion.}   On one hand, the texture must possess strong adversarial properties in the DNN feature space to deceive machine detectors. On the other hand, it needs to achieve high visual naturalness to evade human scrutiny and blend into the environment.

To overcome these challenges, we propose VFACamou, a framework that integrates environmental adaptation with multi-view robustness. VFACamou first ensures visual consistency through an illumination and color estimation module, which extracts and aligns background lighting and dominant colors to achieve natural scene blending. It further handles scale and poses variations via a dynamic multi-scale training strategy, simulating diverse UAV observation conditions to maintain effectiveness across viewpoints. In addition, 
VFACamou introduces a camouflage-specialized loss function that optimizes color distribution matching in a dedicated color space, jointly trained with an adversarial detection loss to balance stealth and evasion performance. Experiments on multiple detectors confirm that VFACamou significantly reduces detection rates while preserving environmental coherence, demonstrating a practical approach for physical-world adversarial camouflage.
Our contributions are summarized as follows:
\begin{itemize}
    \item We design a brand-new end-to-end wearable adversarial camouflage generation method that automatically produces adversarial textures for UAV physical observation environments.
    \item We propose an IoU-guided detection-evasion loss and a color-pattern adaptation loss, which significantly reduce detection rates on multiple mainstream detectors while maintaining texture naturalness.
    \item Our extensive experiments on eight object detectors demonstrate that VFACamou achieves robust physical attack performance while maintaining high perceptual naturalness, surpassing existing methods. 
\end{itemize}

\section{Related Works}
\subsection{Object Detectors for Pedestrian Detection}
With the rapid deployment of deep learning–based object detectors in safety-critical applications such as surveillance and autonomous driving, pedestrian detection has become one of the most important and widely studied detection tasks.
Modern pedestrian detectors are typically built upon general-purpose object detection frameworks, including two-stage detectors such as Faster R-CNN\cite{ren2015fasterrcnn}, one-stage detectors such as the YOLO series\cite{redmon2018yolov3}.
These detectors achieve strong performance by learning robust visual representations of human appearance across large-scale datasets.
In parallel, researchers have increasingly investigated adversarial attacks against object detectors.
However, the vulnerability of pedestrian detectors to adversarial perturbations, especially in the physical world, remains an open and critical challenge.

\subsection{Adversarial Attacks on Human Bodies}
Recent studies have explored adversarial attacks~\cite{song2026erosion,yan2026transferable,zhu2026transferable} that specifically target human-related detection scenarios.
While digital-domain attacks can directly manipulate input pixels, physical-world adversarial attacks must remain effective under varying illumination, viewpoints, material properties, and geometric transformations, making them substantially more challenging \cite{thys2019fooling,zhou2023downstream,zhou2025numbod}.
In pedestrian detection scenarios, this challenge is further exacerbated by the highly dynamic and non-rigid nature of human targets.
Most existing physical attacks are designed for rigid or weakly deformable objects \cite{thys2019fooling} and therefore generalize poorly to humans, whose appearance varies significantly across poses, clothing, and viewpoints.
Although recent studies explore 3D reconstruction, multi-view constraints, or texture-based attacks to improve robustness \cite{Li2023Adv3DG3,hu2023physically}, they typically rely on fixed-topology models and fail to adequately capture human dynamics and non-rigid deformations, resulting in limited adaptability under real-world conditions.
\section{Methodology}\label{sec:methodology}

\begin{figure*}[t]
\centering \includegraphics[width=0.8 \textwidth]{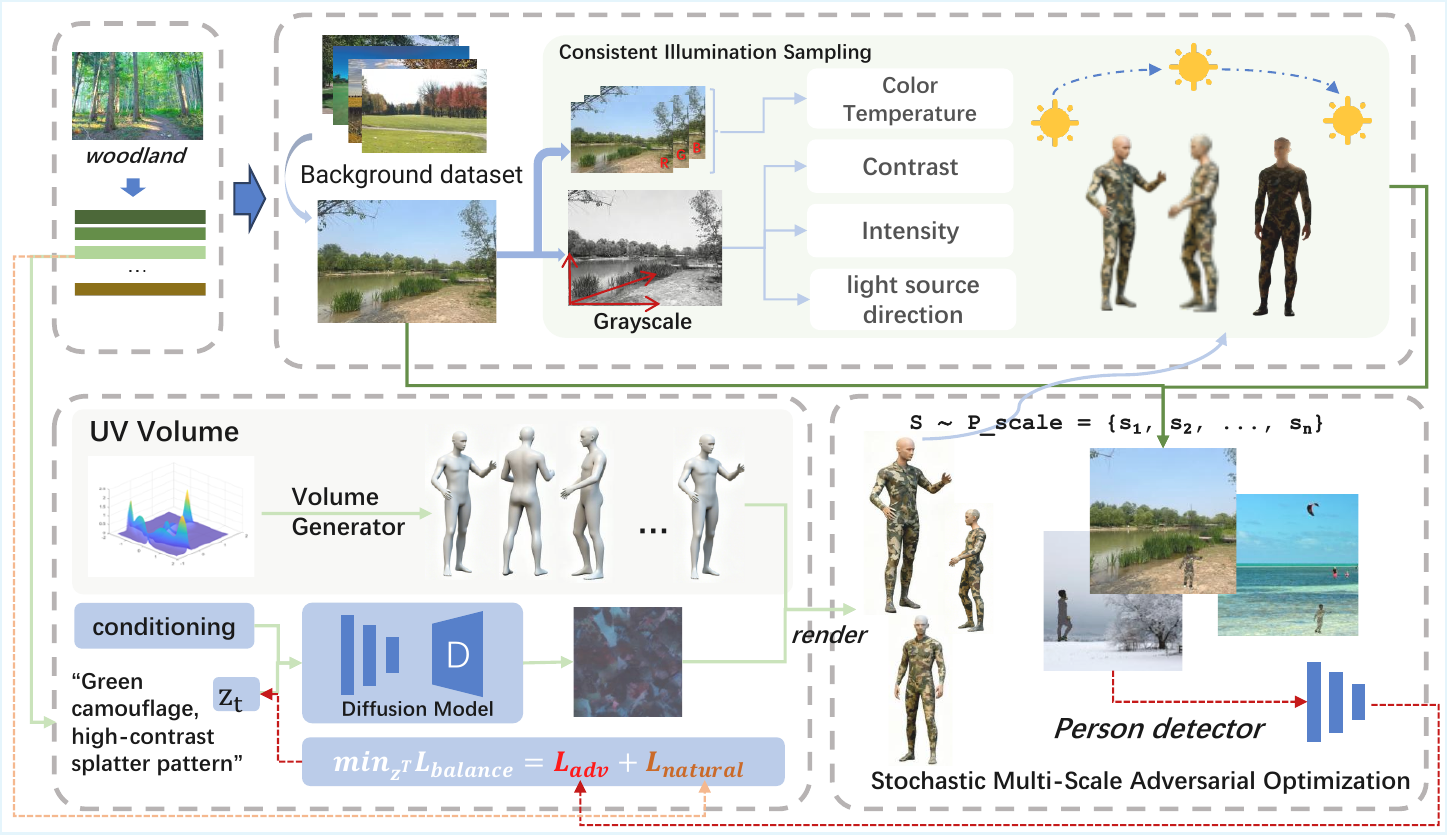}
\caption{An overview of our proposed method.}
\label{fig:pip} 
\end{figure*}

\subsection{Problem Definition}
The core objective of VFACamou is to design a wearable garment pattern that, when captured by a visual sensor, can systematically deceive a pre-trained pedestrian detector. 
Let $B_{pred} = \{b_1, b_2, \ldots, b_n\}$ denote the set of predicted bounding boxes generated by the target detector for input image $I_{adv}$, where each bounding box $b_i$ contains coordinates $(x_i, y_i, w_i, h_i)$ and class confidence $c_i$. $B_{gt} = \{g_1, g_2, \ldots, g_m\}$ represents the set of ground truth human bounding boxes.

\subsection{Motivation}

Existing adversarial attacks against object detectors typically achieve evasion by directly reducing the confidence of the “person” class. However, for humans as non-rigid targets, lowering classification confidence alone does not ensure effective evasion: the detector may still output accurately localized bounding boxes, which reveal the presence of the target.

We argue that under UAV-based physical observation, effective adversarial attacks on humans should disrupt the detector’s overall spatial perception of the target rather than merely weakening its classification confidence. 
Specifically, successful evasion satisfies at least one of the following conditions:

\begin{itemize}
\item \textbf{Target disappearance:} the detector produces no high-confidence predictions near the true human region.
\item \textbf{Target misidentification:} predictions that overlap with the true human region are misclassified as non-“person” categories.
\end{itemize}

Following this intuition, we design an IoU-guided detection-evasion loss that directly constrains the spatial overlap between predicted bounding boxes and ground-truth human boxes, thereby effectively disrupting the detector’s output structure.

\subsection{Multi-Objective Hybrid Optimization Framework}
We formulate adversarial texture generation as a multi-objective optimization problem. The overall loss is defined as:

\begin{equation}
\mathcal{L}_{total} = \mathcal{L}_{det} + \lambda_{nat}\mathcal{L}_{nat}
\end{equation}
where ${L}_{det}$ is the maximum IoU Loss, ${L}_{nat}$ is the natural-environment adaptation loss, $\lambda_{nat}$ is a balancing coefficient controlling the trade-off between attack strength and naturalness.

In UAV surveillance, a target's appearance can vary significantly due to changes in viewing angle, distance, and lighting. To maintain stable adversarial performance under these dynamic conditions, we construct a 3D environment augmentation pipeline that enhances robustness across three dimensions: viewpoint, scale, and illumination.

\noindent\textbf{Viewpoint Randomization.} To account for the variability in UAV trajectories and viewing directions, we randomly sample azimuth and elevation angles during training:
$\theta_{azim} \sim \mathcal{U}(-180^\circ, 180^\circ)$, $\theta_{elev} \sim \mathcal{U}(2^\circ, 18^\circ)$.
This strategy forces the generator to learn viewpoint-invariant adversarial patterns, thereby maintaining attack effectiveness from diverse UAV positions and mitigating overfitting common in fixed-viewpoint rendering.

\noindent\textbf{Multi-Scale Human Rendering}
To account for scale variations caused by changes in UAV altitude, we employ a discretized multi-scale sampling scheme:
\begin{equation}
s \sim \mathcal{C}([0.3, 0.35, 0.4, 0.45]), \quad H_{scaled} = s \cdot H_{original}.
\end{equation}

This not only simulates the continuous size changes of a human in UAV imagery but also guides the generator to develop scale-aware texture patterns. 

\noindent\textbf{Illumination-Consistent Rendering.} To enforce illumination realism, we integrate physically based rendering with an illumination estimation module. A pretrained estimator $f_{\text{light}}(\cdot)$ infers global lighting parameters—such as ambient intensity, primary light direction, and color—from the background image.
During rendering, the 3D human model $M$ with adversarial texture $T{adv}$ is placed under illumination $L_{estimated}$ to produce an image consistent with the background lighting:
\begin{equation}
I_{lit} = \text{Render}_{\text{consistent}}(M, T_{adv}, L_{estimated})
\end{equation}
\begin{equation}
L_{estimated} = f_{\text{light}}(I_{background})
\end{equation}

\subsubsection{Maximum IoU Loss}

In the overall loss of the aforementioned multi-objective optimization problem, the Maximum IoU-based loss function is defined as:

\begin{equation}
\mathcal{L}_{det} = \max_{i \in [1,n]} \left[ \text{IoU}_{ij} \cdot \mathbb{I}(c_i > \tau_{conf}) \cdot \mathbb{I}(\text{cls}_i = \text{person}) \right]
\end{equation}
where $\tau_{conf}$ is the confidence threshold.
$\mathbb{I}(\text{cls}_i = \text{person})$ is a class indicator function.
$\mathbb{I}(c_i > \tau_{conf})$ ensures only predictions above the confidence threshold are considered.

\subsubsection{Natural-Environment Adaptation Loss}
While the generated adversarial textures can mislead detectors, their high-frequency artifacts and unnatural colors make them easily identifiable in real environments. Therefore, we introduce a natural-environment adaptation loss.
Color serves as the primary factor in visual camouflage. We encode common military color schemes—such as jungle green, desert yellow, and urban gray—into a reference color set ${C_{military}^k}_{k=1}^K$. The color adaptation loss is defined as:

\begin{equation}
\mathcal{L}_{color} = \sum_{k=1}^{K} w_k \cdot \min \| C_{tex} - C_{military}^k \|_2^2
\end{equation}
where $w_k$ is a dynamically adjusted weight according to the target deployment environment. By minimizing the distance between the generated texture and the reference color schemes, this loss enforces global tonal consistency with the surrounding background. The minimum-distance formulation allows different regions of the texture to align with different dominant colors, thereby capturing the natural color variations commonly observed in real-world environments.

In addition, observations from military-pattern design show that effective camouflage typically exhibits moderate local contrast and multi-scale irregularity. So we introduce local variance as a measure of pattern complexity. The pattern-control loss is defined as:
\begin{equation}
\mathcal{L}_{pattern} = \mathbb{E}_{\text{scale}} \left[ \text{Var}_{\text{local}}(T) - \sigma_{target}^2 \right]^2
\end{equation}
where $\text{Var}_{\text{local}}(T)$ computes the local variance of the texture $T$ within sliding windows, $\sigma{target}^2$ denotes the target variance level, and $\mathbb{E}_{\text{scale}}$ represents the expectation over multiple image scales.
Combining the color-adaptation and pattern-control objectives yields the final natural-environment adaptation loss:

\begin{equation}
\mathcal{L}_{nat} = \mathcal{L}_{color} + \mathcal{L}_{pattern}
\end{equation}
where $\mathcal{L}_{color}$ ensures texture colors align with military camouflage palette.
$\mathcal{L}_{pattern}$ controls local texture pattern variance within desired ranges.
\section{Experiments}\label{sec:experiments}

\begin{table*}[t!]
\setlength{\abovecaptionskip}{4pt}
\definecolor{gray}{RGB}{240,240,240}
  \centering
    \caption{Attack performance of various methods against different object detectors.}
      \scalebox{0.85}{
    \begin{tabular}{>{\centering\arraybackslash}m{2.0cm}>{\centering\arraybackslash}m{1.8cm}>{\centering\arraybackslash}m{1.8cm}>{\centering\arraybackslash}m{1.8cm}>{\centering\arraybackslash}m{1.8cm}>{\centering\arraybackslash}m{1.8cm}>{\centering\arraybackslash}m{1.8cm}>{\centering\arraybackslash}m{1.8cm}>{\centering\arraybackslash}m{1.8cm}>{\centering\arraybackslash}m{1.8cm}}
    \toprule[1.5pt]
    \multirow{2}[4]{*}{Method} & \multicolumn{8}{c}{Victim Models} \\
\cmidrule(lr){2-9}  & FRCNN & YOLOv8 & MRCNN & Retina & SSD & YOLOv3 & YOLOv9 & FCOS \\
    \midrule
    AdvYolo & \underline{2.70} &47.78  & 1.28 &6.14  &36.56  & 3.82 & 0.12 &5.94  \\
    AdvTexture & \underline{12.08} & 39.42 & 8.18 & 41.52 & 37.30 & 12.06 & 12.20 & 19.20 \\
    AdvCamou & \underline{3.64} & 15.80 & 2.46 & 7.82 & 31.38 & 17.66 & 0.08 & 5.98 \\
    UV-Attack & \underline{43.28} & 36.96 & 52.42 & 45.94 & 48.66 & 11.86 & 5.08 & 62.18 \\
    \cellcolor{gray}Ours & \cellcolor{gray}\underline{93.02} & \cellcolor{gray}66.10 & \cellcolor{gray}95.60 & \cellcolor{gray}89.72 & \cellcolor{gray}78.40 & \cellcolor{gray}40.32 & \cellcolor{gray}56.34 & \cellcolor{gray}95.60 \\
    \midrule
    AdvYolo & 2.40 & \underline{28.38} & 1.54 & 6.66 & 32.38 & 15.52 & 0.14 & 5.26 \\
    AdvTexture &3.14  &\underline{26.58}  &1.08  &7.82  &24.78  &5.20  &0.22  & 5.58 \\
    AdvCamou & 2.42 & \underline{11.36} & 1.54 & 5.48 & 26.58 & 5.88 & 0.12 & 4.06 \\
    UV-Attack & 52.00 & \underline{58.36} & 72.02 & 62.14 & 60.86 & 35.88 & 21.90 & 70.06 \\
    \cellcolor{gray}Ours & \cellcolor{gray}53.74 & \cellcolor{gray}\underline{66.86} & \cellcolor{gray}78.72 & \cellcolor{gray}57.96 & \cellcolor{gray}57.14 & \cellcolor{gray}24.56 & \cellcolor{gray}13.64 & \cellcolor{gray}68.46 \\
    \bottomrule[1.5pt]
    \end{tabular}%
    }
  \label{tab:attack_transferability}%
\end{table*}%

\begin{table*}[t!]
\setlength{\abovecaptionskip}{4pt}
\definecolor{gray}{RGB}{240,240,240}
  \centering
    \caption{Comparison of different methods in terms of ASR (\%) on Faster R-CNN and YOLOv8 ($\tau_{\text{conf}}=0.5$).}
      \scalebox{0.85}{
    \begin{tabular}{>{\centering\arraybackslash}m{2.0cm}>{\centering\arraybackslash}m{1.8cm}>{\centering\arraybackslash}m{1.8cm}>{\centering\arraybackslash}m{1.8cm}>{\centering\arraybackslash}m{1.8cm}>{\centering\arraybackslash}m{1.8cm}>{\centering\arraybackslash}m{1.8cm}>{\centering\arraybackslash}m{1.8cm}>{\centering\arraybackslash}m{1.8cm}>{\centering\arraybackslash}m{1.8cm}}
    \toprule[1.5pt]
    \multirow{2}[4]{*}{\begin{tabular}{c}Method\\\end{tabular}} & \multicolumn{4}{c}{Faster R-CNN} & \multicolumn{4}{c}{YOLOv8} \\
\cmidrule(lr){2-5}\cmidrule(lr){6-9}
    & $\text{IoU}_{0.01}$ & $\text{IoU}_{0.1}$ & $\text{IoU}_{0.3}$ & $\text{IoU}_{0.5}$ 
& $\text{IoU}_{0.01}$ & $\text{IoU}_{0.1}$ & $\text{IoU}_{0.3}$ & $\text{IoU}_{0.5}$ \\
    \midrule
    AdvYolo & 2.70 & 2.70 &3.50  & 2.72 & 29.38 & 28.38 & 29.38 & 30.42 \\
    AdvTexture & 8.12 & 12.08 & 37.22 & 62.20 &28.06  & 26.58 & 27.78 & 26.76 \\
    AdvCamou & 3.64 & 3.64 & 3.50 & 3.68 & 11.74 & 11.36 & 12.04 & 12.08 \\
    UV-Attack & 43.96 & 43.28 & 45.42 & 46.60 & 58.12 & 58.36 & 57.50 & 57.58 \\
    \cellcolor{gray}Ours & \cellcolor{gray}93.10 & \cellcolor{gray}93.02 & \cellcolor{gray}93.34 & \cellcolor{gray}93.18 & \cellcolor{gray}66.82 & \cellcolor{gray}66.86 & \cellcolor{gray}68.76 & \cellcolor{gray}66.84 \\
    \bottomrule[1.5pt]
    \end{tabular}%
    }
  \label{tab:comparison}%
\end{table*}%

\subsection{Experimental Setup}
\noindent\textbf{Datasets and models.}
For a comprehensive evaluation, we use ZJU-Mocap~\cite{peng2021neural} dataset. Specifically, we select the following models: Faster R-CNN~\cite{girshick2015fast}, Mask R-CNN~\cite{he2018mask}, RetinaNet~\cite{lin2018focal}, SSD~\cite{liu2016ssd}, YOLOv3~\cite{redmon2018yolov3}, YOLOv8~\cite{Jocher_Ultralytics_YOLO_2023}, YOLOv9~\cite{wang2024yolov9}, and FCOS~\cite{tian2019fcos}. 

\noindent\textbf{Evaluation metrics.}
We adopt Attack Success Rate (ASR) as the evaluation metric to measure attack effectiveness. ASR is defined as the proportion of instances in which the target person is not successfully detected by the detector. To provide a comprehensive evaluation, ASR is reported under multiple IoU thresholds ranging from 0.01 to 0.5, where IoU represents the overlap between detection bounding boxes and ground truth annotations.

\noindent\textbf{Implementation details.}
The training process runs for 300 epochs with a batch size of 4, and all optimization and inference procedures are conducted on a single NVIDIA A100 GPU. Our adversarial textures are trained and evaluated under multi-pose conditions, where each sample contains diverse human poses and viewing angles to reflect realistic human motion scenarios.

\begin{table}[t!]
\setlength{\abovecaptionskip}{4pt}
\definecolor{gray}{RGB}{245,245,245}
  \centering
    \caption{Ablation study. The Impact of natural-environment adaptation loss on VFACamou.}
      \scalebox{0.95}{
    \begin{tabular}{>{\centering\arraybackslash}m{2.5cm}>{\centering\arraybackslash}m{2.0cm}>{\centering\arraybackslash}m{2.0cm}}
    \toprule[1.5pt]
    Metric & w/ $\mathcal{L}_{\text{nat}}$ & w/o $\mathcal{L}_{\text{nat}}$ \\
    \midrule
    Color Similarity (\%) & 75.32 & 64.22 \\
    \bottomrule[1.5pt]
    \end{tabular}%
    }
  \label{tab:ablation_lnat}%
\end{table}%

\begin{table}[t!]
\setlength{\abovecaptionskip}{4pt}
\definecolor{gray}{RGB}{245,245,245}
  \centering
    \caption{Ablation study. The Impact of multi-scale dynamic training on VFACamou ($\tau_{\text{IoU}}=0.1$, $\tau_{\text{conf}}=0.5$). }
      \scalebox{0.95}{
    \begin{tabular}{>{\centering\arraybackslash}m{2.8cm}>{\centering\arraybackslash}m{2.0cm}>{\centering\arraybackslash}m{2.0cm}}
    \toprule[1.5pt]
    Metric & w/ MDT & w/o MDT \\
    \midrule
    ASR (\%)  & 93.02 & 58.96 \\
    \bottomrule[1.5pt]
    \end{tabular}%
    }
  \label{tab:ablation_mdt}%
\end{table}%

\subsection{Digital Attack Performance}

We conduct digital experiments on diverse person detectors to evaluate the effectiveness and transferability of the proposed attack under multi-pose scenarios. Adversarial textures are optimized on Faster R-CNN and YOLOv8, and subsequently transferred to unseen models. Performance is measured by the Attack Success Rate (ASR) with an IoU threshold of 0.1 and a confidence threshold of 0.5. We compare against representative texture- and patch-based methods, including AdvCamou~\cite{hu2023physically}, UV-Attack~\cite{li2025uv}, AdvTexture~\cite{hu2022adversarial}, and AdvYOLO~\cite{thys2019fooling}. For fairness, patch-based methods are tiled over the texture surface.

As shown in \cref{tab:attack_transferability}, the proposed method consistently outperforms all baselines in both white-box and cross-model transfer settings. When trained on Faster R-CNN, it achieves strong performance on the source model while maintaining high transferability across heterogeneous detectors. Similarly, training on YOLOv8 yields superior attack effectiveness, surpassing competing approaches. Overall, the results demonstrate robust digital attack performance and strong cross-model transferability in multi-pose scenarios.

\subsection{Comparison Study}

To further assess the competitiveness of our method, we perform a comparative study against existing adversarial person-detection attacks under consistent multi-pose training and testing settings, including AdvCamou~\cite{hu2023physically}, UV-Attack~\cite{li2025uv}, AdvTexture~\cite{hu2022adversarial}, and AdvYOLO~\cite{thys2019fooling}. 
The experiments are conducted on Faster R-CNN and YOLOv8 as victim detectors. The ASR is evaluated under multiple IoU thresholds (0.01,0.1,0.3,0.5) with a confidence score of 0.5.

As shown in \cref{tab:comparison}, our method achieves the highest ASR across all IoU thresholds on both detectors. On Faster R-CNN, our method demonstrates exceptional performance with ASR consistently above 93\% across all IoU settings, significantly outperforming all baseline methods. On YOLOv8, our method also achieves better attack performance than other methods. Notably, our method shows relatively stable performance across different IoU thresholds, indicating that our generated adversarial textures achieve more robust and consistent attack performance.

\subsection{Physical Attack Performance}
\begin{figure}[t]
\centering \includegraphics[width=0.48 \textwidth]{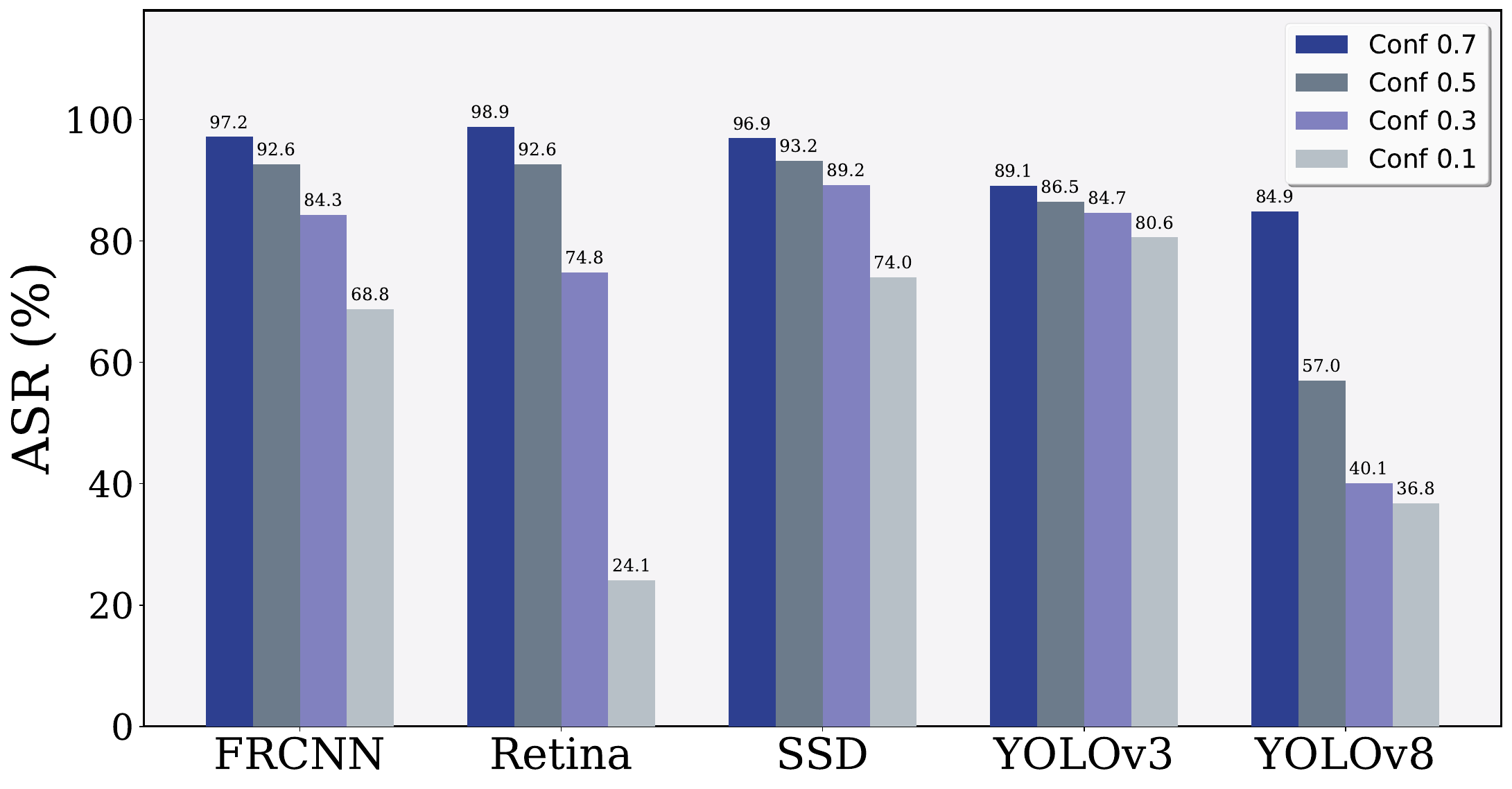}
\caption{Physical attack success rates across different detectors and confidence thresholds.}
\label{fig:physical} 
\end{figure}

We evaluate the real-world effectiveness of the proposed attack by conducting physical experiments under multi-pose, multi-view, and varying-distance outdoor conditions.
The adversarial textures trained on Faster R-CNN are printed on a long-sleeve T-shirt and trousers, and a subject wearing the garments is recorded using an aerial drone. The collected video sequence contains continuous variations in posture, camera viewpoint, and subject-to-camera distance.
Each sequence is then evaluated on multiple detectors, and a frame is regarded as successfully attacked when the person is not detected in that frame. 
$ASR = 1-N_{detected}/N_{total}$,
where $N_{detected}$ denotes the number of frames in which the person is detected, and $N_{total}$ denotes the total number of frames in the sequence.
The ASR results on Faster R-CNN, RetinaNet, SSD, YOLOv3, and YOLOv8 under different confidence thresholds are shown in \cref{fig:physical}. These results demonstrate that the generated textures maintain strong adversarial capability in real-world environments and remain effective under changes in pose, viewing angle and distance.

\subsection{Ablation Study}
In this section, we analyze the contribution of the key components in our framework. All ablation experiments are conducted on the Faster R-CNN detector under the same training and evaluation settings as the main experiments.

\noindent\textbf{The effect of natural-environment adaptation loss.}
We first investigate the influence of incorporating the natural-environment adaptation loss during optimization. This loss encourages adversarial textures to better align with real-world illumination and color distributions. We compute color similarity values composed of histogram intersection similarity and color distance similarity with and without $L_{nat}$, as shown in \cref{tab:ablation_lnat}, demonstrating that the textures maintain closer visual consistency with the surrounding environment. 

\noindent\textbf{The effect of multi-scale dynamic training.}
We further evaluate the role of the multi-scale dynamic training strategy, which simulates variations in observation distance and human scale during training. This strategy enables the adversarial textures to remain effective under diverse viewing conditions. The results in \cref{tab:ablation_mdt} show that enabling multi-scale dy namic training significantly improves the attack success rate.

\section{Conclusion}\label{sec:conclusion}

In this paper, we propose VFACamou, an end-to-end framework for generating wearable adversarial camouflage that remains effective under UAV reconnaissance conditions, including dynamic viewpoints, body deformation, and complex illumination. VFACamou combines UV-volume rendering with a diffusion-based texture generator to ensure geometry-consistent appearance across poses and scales, while an illumination–color consistency estimator facilitates adaptive texture adjustment for realistic environmental blending. To enhance robustness, a multi-scale dynamic training strategy is employed to simulate UAV-induced variations in distance, viewpoint, and articulation. Extensive digital and physical experiments demonstrate that VFACamou consistently outperforms existing methods, achieving superior attack performance across multiple mainstream detectors.

\bibliographystyle{IEEEtran}
\bibliography{ref}

\end{document}